# Online high-precision prediction method for injection molding product weight by integrating time series/non-time series mixed features and feature attention mechanism

Maoyuan Li[a,c], Sihong Li[c], Guancheng Shen*[b,c], Yun Zhang*[c], Huamin Zhou[c]

* Corresponding author: marblezy@hust.edu.cn (Yun Zhang); gc_shen@outlook.com (Guancheng Shen)

[a.] School of Aerospace Engineering, Xiamen University, Xiamen, 361005, China;

[b.] Xi'an Modern Chemistry Research Institute, Xi'an 710065, China;

[c.] State Key Laboratory of Materials Processing and Die & Mould Technology, School of Materials Science and Engineering, Huazhong University of Science and Technology, Wuhan 430074, China.

## Abstract

To address the challenges of untimely detection and online monitoring lag in injection molding quality anomalies, this study proposes a mixed feature attention-artificial neural network (MFA-ANN) model for high-precision online prediction of product weight. By integrating mechanism-based with data-driven analysis, the proposed architecture decouples time series data (e.g., melt flow dynamics, thermal profiles) from non-time series data (e.g., mold features, pressure settings), enabling hierarchical feature extraction. A self-attention mechanism is strategically embedded during cross-domain feature fusion to dynamically calibrate inter-modality feature weights, thereby emphasizing critical determinants of weight variability. The results demonstrate that the MFA-ANN model achieves a RMSE of 0.0281 with ± 0.5 g weight fluctuation tolerance, outperforming conventional benchmarks: a 25.1% accuracy improvement over non-time series ANN models, 23.0% over LSTM networks, 25.7% over SVR, and 15.6% over RF models, respectively. Ablation studies quantitatively validate the synergistic enhancement derived from the integration of mixed feature modeling (contributing 22.4%) and the attention mechanism (contributing 11.2%), significantly enhancing the model's adaptability to varying working conditions and its resistance to noise. Moreover, critical sensitivity analyses further reveal that data resolution significantly impacts prediction reliability, low-fidelity sensor inputs degrade performance by 23.8% RMSE compared to high-precision measurements. Overall, this study provides an efficient and reliable solution for the intelligent quality control of injection molding processes.

**1. Introduction**

Plastic products are extensively used in high-end manufacturing fields such as national defense, military, and aerospace due to their advantages of lightweight, high specific strength, excellent optical properties, and low cost [1]. Injection molding remains the dominant production technique for these parts, representing roughly 80% of all polymer component fabrication [2]. However, in highly automated and high-throughput injection molding environments, existing inspection methods fail to keep pace with the rapid cycle times, leading to significant bottlenecks in quality assurance [3]. For example, although the cycle time for a single precision optical lens may be as short as 10 s, the ensuing multi-dimensional inspection process, including dimensional metrology, weight verification and surface-quality assessment, can demand 5-10 min per piece [4]. This stark discrepancy between detection efficiency and production rate represents a critical challenge that must be addressed to achieve both efficient and accurate quality inspection.

Quality monitoring of injection molding products has traditionally relied on offline sampling inspections, which come with several drawbacks: high testing costs, long testing cycles, poor real-time performance, and a tendency to overlook abnormalities within batches [5]. As plastic products increasingly demand higher quality and the process windows narrow, rising defect rates render offline sampling insufficient for real-time monitoring. In contrast, data-driven online prediction methods construct a mapping model between process parameters and quality indicators [6], offering high computational efficiency and real-time prediction capabilities. For instance, Engel Co., Ltd., has developed the iQ weight control system [7], which achieves product weight stability through real-time regulation of injection parameters. Existing weight prediction modeling methods for injection molding products can generally be classified into two categories: 1) statistical learning and classical machine learning methods, such as partial least squares regression (PLS) [8], support vector regression (SVR) [9], and random forest (RF) [10]; 2) deep learning methods [11-13], including artificial neural networks (ANN), ensemble learning, etc. For example, Chen et al. [14] employed variables like barrel temperature, back pressure, maximum mold surface temperature, and injection pressure integral as inputs to develop an ANN model for predicting the weight of injection-molded products. Similarly, Shi et al. [15] proposed a product quality prediction model for injection molding based on temperature field infrared thermography and convolutional neural network. However, these studies generally overlook the time series dependencies inherent in injection molding process data, leading to insufficient extraction of dynamic feature information.

Traditional machine learning methods typically rely on manual feature engineering to extract time series information and improve model accuracy. For instance, Yuan et al. [16] utilized linear dynamic system modeling to extract time series features in industrial processes, thereby enhancing the accuracy of soft measurement models. Recurrent neural network (RNN) inherently captures time dependencies through their sequential modeling capabilities, significantly reducing the need for manual feature engineering. Enhanced RNN architectures like long short-term memory (LSTM) network and gated recurrent unit (GRU) mitigate the gradient vanishing problem via gating mechanisms, performing effectively in long-sequence modeling tasks. For example, Gao et al. [17] constructed a GRU-based autoencoder with an attention mechanism model, which significantly improves part weight prediction accuracy by deeply mining time series features from sensor waveforms. Similarly, Guo et al. [18] proposed a cavity pressure feature extraction method based on variational autoencoders (VAE), which greatly optimized 3D model retrieval efficiency. However, these methods mainly focus on extracting features from waveform data within a single mold, and do not fully account for the dependencies among features from different molds that can affect injection molding product weight prediction. For example, Ketonen et al. [19] proposed the variational autoencoder-based anomaly detector (DSVAE-AD) model, which detects data anomalies and provides process insights by considering the feature dependencies between different molds in the injection molding process. It is important to note, though, that not all injection molding process features exhibit clear time series characteristics across different molds. Blindly applying time series dependencies can sometimes introduce noise interference, potentially degrading the overall model performance.

Some studies [20, 21] indicate that while time series features capture the dynamic changes in data, non-time series features provide valuable global context. By constructing a mixed feature prediction model that incorporates both time series and non-time series features, the relationship between input variables and target outcomes can be more comprehensively described, enhancing both the accuracy and robustness of the prediction model. The selection of input variables is crucial to the performance of the prediction model for injection molding product weight. Existing studies often rely on manual expertise combined with static feature selection methods (such as Filter, Wrapper, and Embedded approaches) to optimize model performance. For example, Sun et al. [22] employed a filtering feature selection method based on Copula information entropy to significantly improve the dimensional prediction accuracy of injection molding products. Liu et al. [23] introduced sparsity constraints via partial least squares regression to select key variables that contribute most to the target variable, thereby enhancing both the prediction performance and interpretability of the model. However, the operating condition drift caused by environmental changes and disturbances during the injection molding process leads to the dynamic evolution of feature importance. The fixed-weight mechanism inherent in static feature selection methods may result in the loss of key information, adversely affecting the accuracy and stability of the prediction model. The attention mechanism [24], which dynamically selects and focuses on relevant parts of an input sequence, has shown significant advantages in industrial monitoring applications, such as gear life prediction[25] and electric field state estimation[26]. While process data-based product weight prediction methods enable online detection, existing studies often overlook the distinctions between time series and non-time series features. Treating all features as time series data can introduce noise, which may degrade the model's performance.

This study proposes a high-precision online prediction method for injection molding product weight by integrating time series/non-time series mixed features and a feature attention mechanism. First, a mixed feature analysis is conducted, combining both the injection molding process mechanism and data-driven insights. The principle, model architecture, parameter settings, and experimental design of the proposed method are then described in detail. Subsequently, the model's prediction accuracy is evaluated against baseline models, ANN and LSTM (using different time series feature inputs), as well as SVR and RF (representing different modeling approaches). The contributions of the mixed feature modeling strategy and the feature attention mechanism to prediction accuracy are also analyzed. The proposed approach significantly enhances both the accuracy and generalization capability of the injection molding product weight prediction model, offering a valuable foundation for intelligent manufacturing, improved product quality, and reduced production costs.

## 2. Mixed feature analysis of injection molding process data
### 2.1 Analysis of mixed characteristics based on injection molding mechanism

During the injection molding process, key characteristic data can be categorized based on execution actions and their specific properties. These data are primarily classified as follows:

Mold characteristic data: his includes mold closing and opening times, which reflect the mold's operational efficiency and cycle control. These data do not exhibit cross-mold time dependence and are considered non-sequential.

Injection holding pressure characteristic data: This includes parameters such as injection peak pressure and actual holding time, which intuitively describe the mold filling state. These data directly affect product weight and are categorized as non-sequential.

Melting characteristic data: This includes parameters such as melting glue time and starting position. Due to the delay and cumulative effects of the melting glue process, the behavior of the current mold can influence the quality of subsequent molds, making these data sequential.

Temperature characteristic data: This includes barrel and mold temperatures, which reflect the periodic temperature cycle along with its delay and cumulative effects. These data impact both the current and subsequent molds, and are considered sequential.

Mold cavity data: Among these, pressure data reflects the current mold filling state and is non-sequential, while temperature data is sequential due to heat conduction and cumulative effects.

The characteristic categories and properties of these injection molding process data are summarized in Table 1.

Table 1 Characteristic classification of injection molding process

| Source | Feature category | Feature properties |
| --- | --- | --- |
| Injection molding machine | Mold features | non-sequential |
|  | Injection pressure holding features | non-sequential |
|  | Melt features | sequential |
|  | Temperature features | sequential |
| Mold cavity | Mold cavity pressure features | non-sequential |
|  | Temperature features | non-sequential |

## 2.2 Data-driven mixed feature analysis

A time series data is a sequence of observations arranged in chronological order, used to record changes in variables over time:

$$X = \{x_1, x_2, ..., x_n\} \qquad (1)$$

As a typical cyclical process, injection molding generates characteristic data that can be expressed as a multivariate time series data. The characteristics of time series data depend on whether the data exhibits dependencies or trends over time.

Autocorrelation analysis helps verify the time dependence between data points by calculating the autocorrelation function (ACF), which checks whether the data value at a certain moment is related to values at previous or future moments. The autocorrelation coefficient at a lag moment is defined as:

$$r_k = \frac{\sum_{t=1}^{n-k}(x_t - \bar{x})(x_{t+k} - \bar{x})}{\sum_{t=1}^{n}(x_t - \bar{x})^2}, k = 0, 1, 2, ... \tag{2}$$

where $k$ is the lag order, $x_t$ and $x_{t+k}$ are the observations at time $t$ and $t+k$ respectively, $\bar{x}$ is the mean of the time series data, and $n$ is the total length of the time series data. The autocorrelation coefficient ranges from -1 to 1. A value close to 1 indicates a strong positive correlation, close to -1 indicates a strong negative correlation, and close to 0 indicates no significant correlation.

The actual switching speed, melting time, injection time, and return water temperature are used as representative non-time series and time series feature data, respectively. The autocorrelation coefficients for these variables are calculated with a 99% confidence interval, which is commonly used to determine whether the autocorrelation coefficient is significant. If the autocorrelation coefficient exceeds the confidence interval, it indicates a significant autocorrelation at that lag, suggesting the presence of time series characteristics. As shown in **Fig. 1**, the autocorrelation coefficients for the actual switching speed and injection time variables are close to zero at all lag orders and do not exceed the set significance limit. This indicates that these variables have no significant correlation with past or future data and do not exhibit time series characteristics. In contrast, the melting time and return water temperature variables exceed the significance limit at the fourth lag, suggesting that they do exhibit notable time series characteristics.

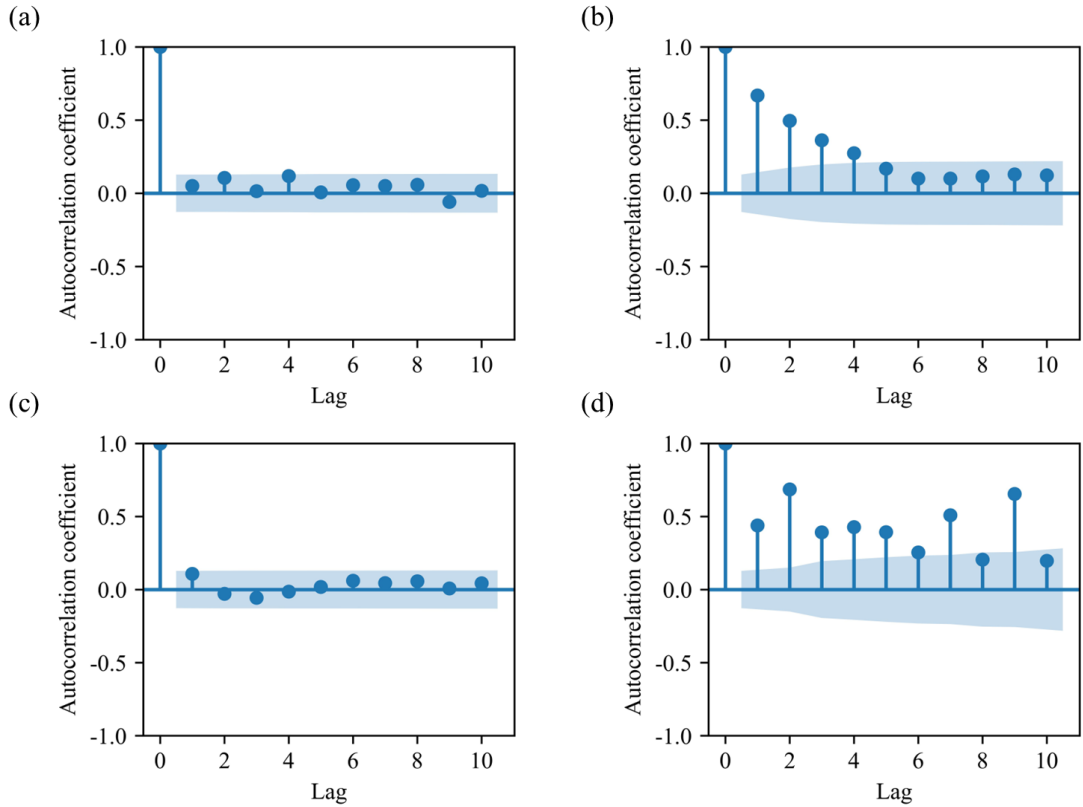

**Fig. 1.** ACF results: (a) actual switching speed; (b) melting time; (c) injection time; (d) return water temperature

## 3. Methodology

### 3.1 Principle of the method

The characteristic data of the injection molding process consists of a mixed feature vector that includes both time series and non-time series features. Effectively modeling these mixed features is essential for improving the accuracy of the product weight prediction model. The LSTM network is a specialized neural network designed to selectively retain or forget information through memory units and gating mechanisms, including forget gates, input gates, and output gates. This unique structure significantly enhances the model's ability to capture time-varying patterns, making LSTM particularly well-suited for modeling the time series characteristics of injection molding time series features and understanding their impact on product weight. **Fig. 2** illustrates a typical LSTM unit structure, where the forget gate $f_t$, input gate $i_i$, and output gate $O_t$ control the flow of information through the network. The calculation process can be expressed as follows:

$$f_t = \sigma\left(W_f \cdot [h_{t-1}, x_t] + b_f\right) \tag{3}$$

$$i_t = \sigma\left(W_i \cdot [h_{t-1}, x_t] + b_i\right) \tag{4}$$

$$o_t = \sigma\left(W_o \cdot [h_{t-1}, x_t] + b_o\right) \tag{5}$$

where $\sigma$ represents the sigmoid activation function, $h_{t-1}$ denotes the hidden state from the previous time step, and $x_t$ is the input at the current time step. $W_f$, $W_i$, and $W_o$ are the weight matrices for the forget gate, input gate, and output gate, respectively, while $b_f$, $b_i$, and $b_o$ are the corresponding bias terms.

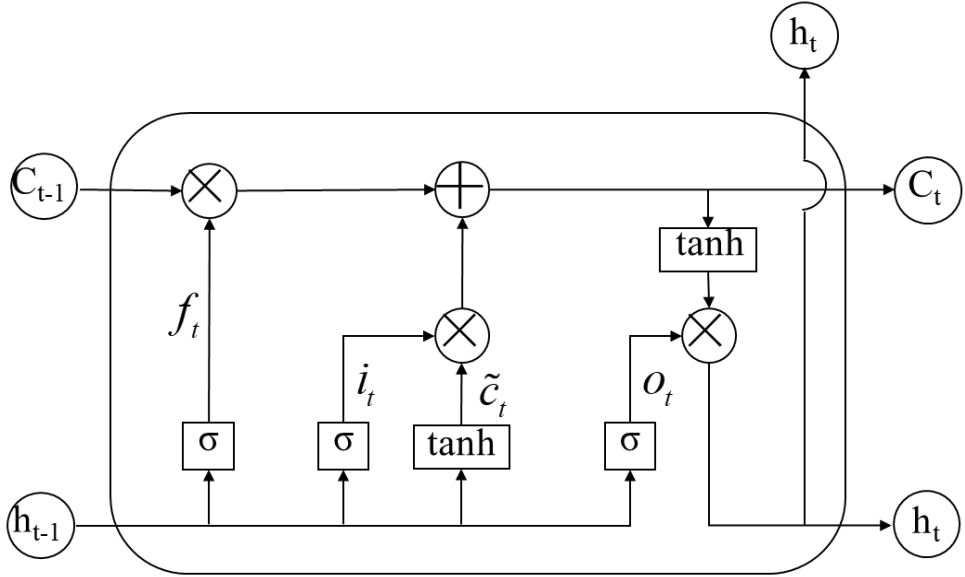

**Fig. 2.** LSTM unit structure

The memory unit updates its internal state through the following calculation process. First, the *tanh* activation function generates candidate information $\tilde{C}_t$. Then, the memory unit state $C_t$ is selectively updated according to the input gate $i_t$. Finally, the updated memory unit state $C_t$ is combined with the output gate to calculate the hidden state $h_t$ of the current time step. The calculation process is as follows:

$$\tilde{C}_t = \tanh\left(W_c \cdot [h_{t-1}, x_t] + b_c\right) \tag{6}$$

$$C_t = f_t * C_{t-1} + i_t * \tilde{C}_t \tag{7}$$

$$h_t = o_t * \tanh(C_t) \tag{8}$$

Through the gating mechanism, LSTM can flexibly and selectively remember and forget information, effectively capturing the time series dependencies in the time series features of the injection molding machine.

The attention mechanism, inspired by human vision and cognitive processes, allows the model to selectively focus on important information while ignoring irrelevant parts. The feature attention mechanism, a variant of the attention mechanism, is specifically designed to dynamically adjust the importance of different feature dimensions. In the injection molding process, some features contribute more significantly to the prediction than others, and changes in working conditions can affect the importance of each feature. The feature attention mechanism adapts by adjusting the weights of the features to emphasize important ones and ignore irrelevant ones, thus preventing the loss of key information. The weight generation layer is the core of the feature attention mechanism, and it reflects the importance of each feature by calculating the attention weight of each feature. This study adopts the self-attention mechanism to generate these weights[24]. Specifically, the input feature matrix is mapped to query, key, and value vectors via linear transformations. The corresponding formulas are as follows:

$$\begin{aligned} Q &= XW_q \\ K &= XW_K \\ V &= XW_v \end{aligned} \tag{9}$$

The similarity between the query vector and the key vector is calculated using the dot product, and the attention weight distribution is then normalized using the SoftMax function. These normalized attention weights are applied to the value vector to obtain the enhanced feature representation. The process can be summarized as follows:

$$\text{Attention}(Q, K, V) = \text{SoftMax}\left(\frac{QK^T}{\sqrt{d_k}}\right)V \qquad (10)$$

where $d_k$ is the dimension of the key vector $K$, which is used to prevent the dot product value from being too large and helps stabilize the gradient.

**3.2 Model architecture**

This study proposes a mixed feature attention-artificial neural network (MFA-ANN) model (**Fig. 3**), aimed at improving the accuracy of injection molding product weight prediction by differentiating between time series and non-time series features. The model utilizes distinct processing techniques to effectively integrate these features within the hidden layers, thereby enhancing prediction performance.

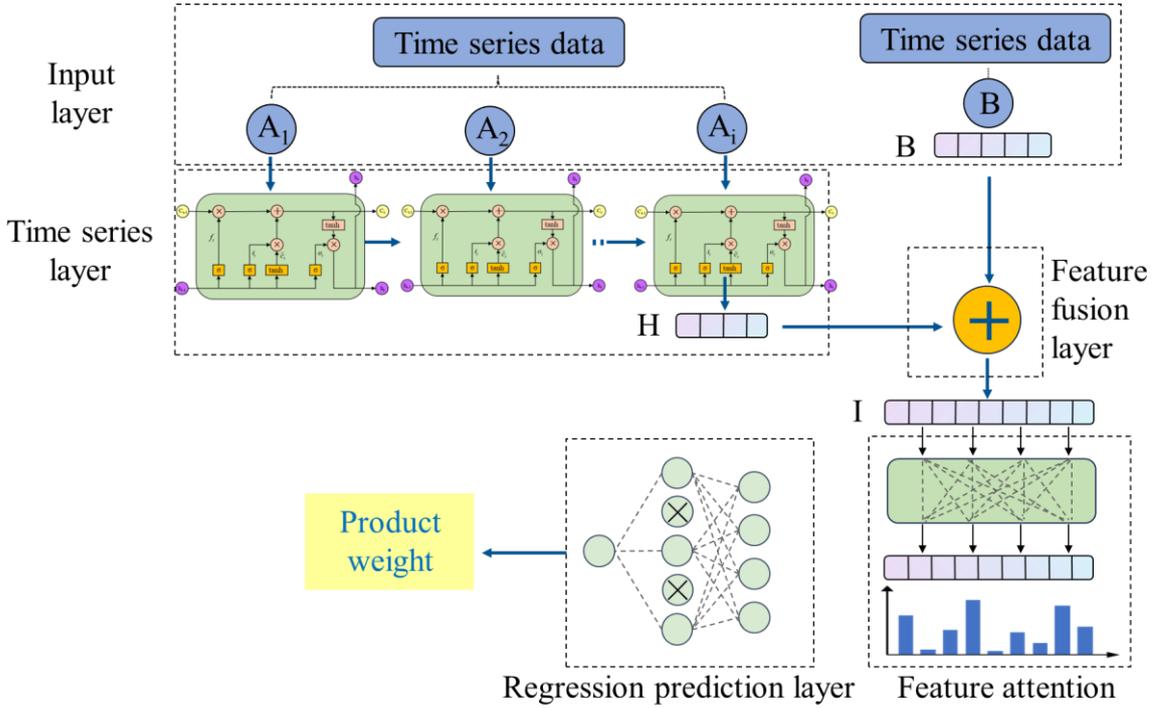

**Fig. 3.** Proposed MFA-ANN model

The MFA-ANN model consists of five components:

1) Input layer: This layer receives both time series feature data (A) and non- time series feature data (B). Time series feature data (A) includes melt features, temperature features, and mold cavity temperature features. Non-time series feature data (B) includes mold features, injection holding pressure parameters, and mold cavity pressure features.

2) Time series processing layer: The time series feature data (A) is passed into an LSTM network, which learns the time series dependencies and generates an abstract representation of the sequence. The hidden state (H) from the last time step of the LSTM aggregates the time dependencies across the entire input sequence, capturing both current and historical module information. This state is then passed to the next layer for regression modeling of the injection molded product weight.

3) Multi-feature fusion layer: This layer concatenates the time series feature representation (the LSTM hidden state H) of the time series feature data (A) with the non-time series feature data (B) to form a mixed feature representation (I), as expressed by the following equation:

$$I = [H; B] \tag{11}$$

4）Feature self-attention layer: The mixed feature representation (I) is fed into the self-attention layer. The model uses the self-attention mechanism to highlight the most important features in the input data, allowing it to focus on key factors and improving the model's prediction accuracy.

5）Prediction layer: After processing by the feature self-attention layer, the model generates an enhanced feature representation. This is then passed to a fully connected regression layer, which outputs a continuous value representing the predicted weight of the injection molded product.

**3.3 Model parameters**

Considering the small size of the injection molding dataset, and to avoid overfitting while enhancing the model's generalization ability, this study designs a simplified network architecture. In the time series processing layer, a single-layer LSTM model is used to process the time series feature data, with the output dimension set to 8, matching the input time series feature dimension. For non-time series feature data, it is directly fed into the multi-feature fusion layer without additional linear transformations, thereby preserving its original physical and process-related information while reducing computational complexity. In the prediction layer, a small fully connected network with a single hidden layer is used, where the ReLU activation function introduces non-linear feature expression. Dropout technology (with a dropout rate of 0.3) is also employed to improve the generalization ability and stability of the model.

During the model training process, the weights and biases between neurons are optimized to progressively capture the characteristics of the injection molding product weight. The loss function is defined as the error between the predicted data and the actual data, and it is expressed as:

$$error = \frac{1}{n}\sum_{1}^{n}(y_i - \hat{y}_i)^2 \qquad (12)$$

where $n=4$ represents the batch size, referring to the number of training samples used in each iteration. The error is updated through backpropagation, and the optimizer used is AdamW, with an initial learning rate of 0.0005. This optimizer retains Adam's dynamic learning rate adjustment and bias correction capabilities, while incorporating a stricter weight decay to help mitigate overfitting.

**3.4 Experiments and online measurement of the injection molding process**

Factors affecting the quality of injection molding products include the stability of molding equipment, changes in raw material properties, set process conditions, manual operations, and fluctuations in the production environment. In large-scale factory production, process personnel typically do not frequently adjust process conditions or intervene manually. The product weight fluctuations are primarily caused by the condition of molding equipment, variations in raw material properties, and environmental fluctuations. To accurately predict product weight changes in a mass production environment, this study collected real production data from 400 consecutive molds, with process conditions held constant and no manual intervention. The data included key features of the injection molding machine, waveform supplement features, and product weight data.

**Fig. 4** shows the injection molding machine, mold, and product used in the experiment. The injection molding machine is a medium-sized horizontal all-electric model produced by Yizumi Co., Ltd., specifically the FF120 model, featuring a screw diameter of 30mm. The plastic material used is polypropylene (PP) from Sinopec Hainan Refining and Chemical Company, brand EP548R. The product is a stirrup, an auxiliary tool commonly used in concrete structure reinforcement and construction projects. In the experiment, the YA-TM6KW-HTW mold temperature controller, also produced by Yizumi Co., Ltd., was used to regulate the mold temperature with a control accuracy of ±0.1°C. The weight of the stirrup product was measured using a Longteng brand JD500-3 precision electronic balance, with a minimum weight resolution of 0.001g. The injection molding process parameters are detailed in Table 2.

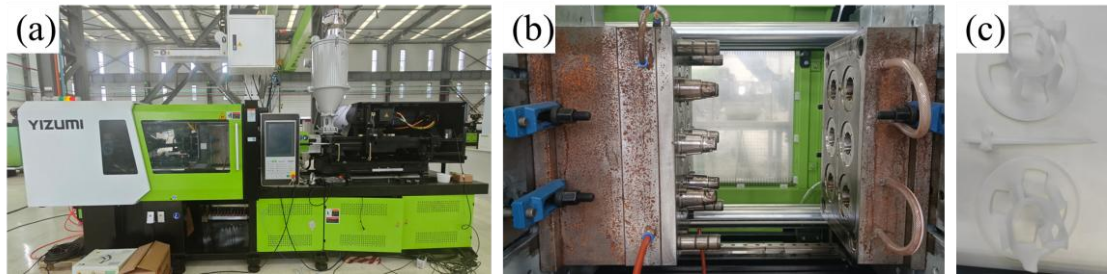

**Fig. 4.** Experimental setup: (a) FF120 injection molding machine; (b) mold; (c) product

Table 2 Experimental process parameters

| Process parameter | Unit | Setting value |
| --- | --- | --- |
| Injection pressure | MPa | 150 |
| Injection speed | mm/s | 150/130/110 |
| Injection position | mm | 70/35 |
| V/P switch position | mm | 15 |
| Holding pressure | MPa | 30 |
| Holding time | s | 0.5 |
| Cooling time | s | 25 |
| Barrel temperature | °C | 200/210/205/175 |
| Mold temperature controller temperature | °C | 45 |

The weight of injection molding products is influenced by numerous factors. By monitoring the process data of molding equipment and the changes in the melt state within the mold cavity, the causes of most product weight fluctuations can be effectively identified. The stability and operational status of the molding equipment, as well as the impact of the set process conditions, can be directly perceived and quantified through the process data. Changes in the raw materials properties can also be reflected in the process data of the molding equipment and the data related to melt state changes in the mold cavity. The influence of manual operations and production environment conditions is primarily transmitted to the molding process through indirect effects on equipment status or material properties, ultimately impacting product weight. Therefore, by comprehensively monitoring the molding equipment's process data and the melt state changes in the mold, reliable data support can be provided for systematic analysis and accurate prediction of product weight fluctuations. **Fig. 5** illustrates the sensor layout of the injection molding system.

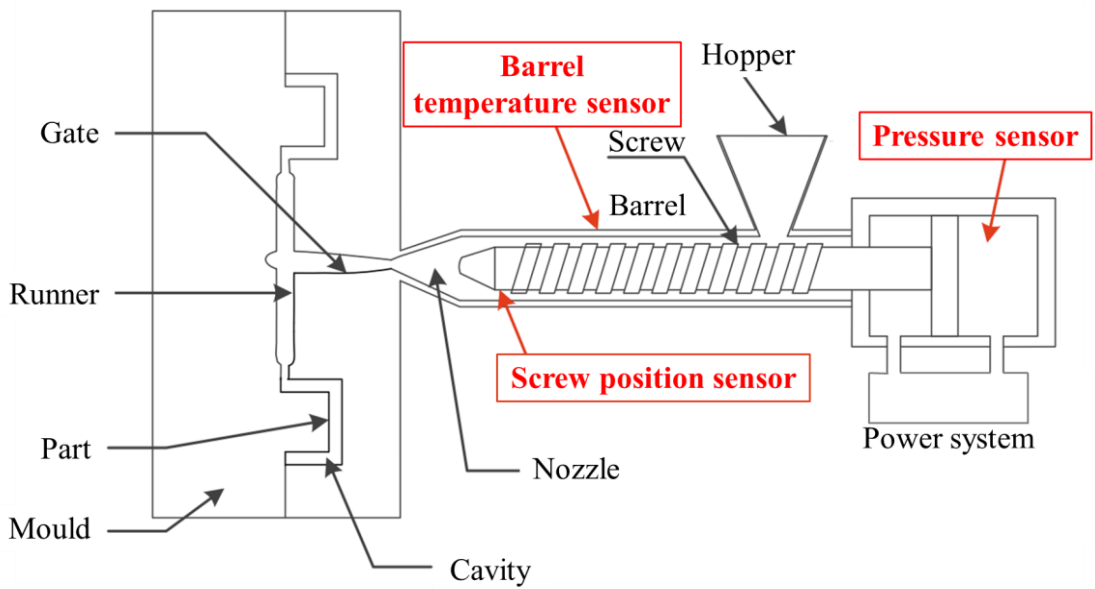

**Fig. 5.** Injection molding process data online measurement system

## 4. Results and discussions

### 4.1 Prediction accuracy analysis of injection molding product weight

The root mean square error (RMSE) is used as a quantitative metric to assess the accuracy of the product weight prediction model. The formula for calculating the

RMSE is:

$$\text{RMSE} = \sqrt{\frac{1}{n}\sum_{i=1}^{n}(y_i - \hat{y}_i)^2} \qquad (13)$$

where $n$ is the number of samples, $y_i$ and $\hat{y}_i$ are the predicted and true values, respectively. In this study, all models are trained using data from the first 100 molds and tested using data from molds 101 to 200 to evaluate prediction accuracy. This data division approach is selected to mitigate the impact of condition drift in the injection molding process, which may cause the model to deviate significantly during long-term predictions. By focusing on short-term prediction accuracy, the effect of condition drift on the test results is minimized, ensuring a more reliable evaluation of the model's performance in static prediction tasks.

**4.1.1 Comparison of prediction accuracy of different feature input methods**

To validate the superiority of the proposed method, this study compares the performance of the MFA-ANN model with two baseline models: an ANN model using non-time series features and an LSTM model using time series features. The hyperparameters of the ANN and LSTM models are consistent with those of the MFA-ANN model, including key parameters such as the number of network layers, the number of nodes, and the learning rate. This ensures that any structural differences between the models do not influence the results, allowing for a fair comparison of how different feature input methods affect prediction accuracy.

As shown in **Fig. 6**, the RMSE of the MFA-ANN model is 0.0281, while the RMSE values for the ANN and LSTM models are 0.0375 and 0.0365, respectively. The MFA-ANN model achieves the lowest RMSE and the highest prediction accuracy. Compared to the two baseline models, the prediction accuracy of the MFA-ANN model is improved by 25.1% and 23.0%, respectively. These results demonstrate that the MFA-ANN model significantly enhances the accuracy and reliability of product weight predictions.

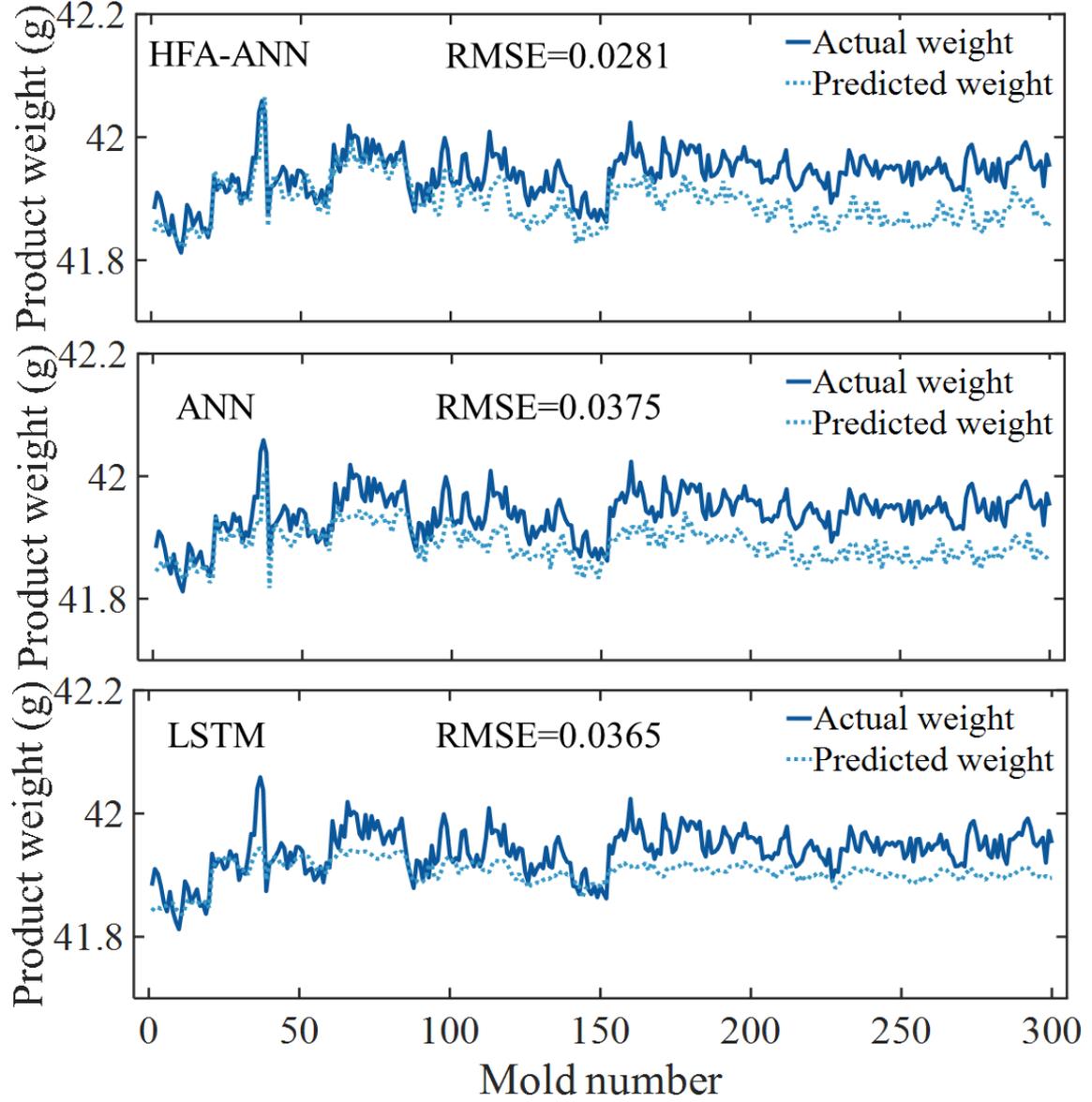

**Fig. 6.** Prediction results of different feature input methods: (a) MFA-ANN model;

(b) ANN model; (c) LSTM model

To further evaluate the prediction performance of the MFA-ANN model, a paired t-test was conducted on the prediction results of all models based on the same test set. The null hypothesis is defined as the mean difference between the two sets of data is 0, that is, $\mu_d = 0$, and the test statistic expression is:

$$t = \frac{\bar{d}}{s_d / \sqrt{n}} \tag{14}$$

where $\bar{d}$ is the mean of the paired sample differences, $s_d$ is the standard deviation of the differences, and *n* is the number of paired samples.

The calculated p-values for the prediction performance differences between the various modeling methods are presented in Table 3. The results show that the p-values between the MFA-ANN model and the baseline models are $4.52 \times 10^{-6}$ and $5.36 \times 10^{-3}$, both of which are below the significance level of 0.05. This indicates that the proposed method significantly outperforms the two baseline models in terms of prediction accuracy. In the comparison between the ANN and LSTM models, the p-value is 0.63, which is greater than 0.05, suggesting that there is no significant difference in prediction performance between the two models.

Table 3 P-values of prediction performance differences of different modeling methods

|         | MFA-ANN | ANN | LSTM |
|---------|---------|-----|------|
| MFA-ANN | × | × | × |
| ANN | $4.52 \times 10^{-6}$ | × | × |
| LSTM | $5.36 \times 10^{-3}$ | 0.63 | × |

The prediction error distribution of the three models is illustrated by the cumulative distribution function (CDF) of the prediction error, as shown in **Fig. 7**. The *x*-axis of the CDF plot represents the prediction error, while the *y*-axis represents the cumulative probability that the error is less than or equal to the corresponding value on the x-axis. The error range of the MFA-ANN model is the smallest, indicating that its errors are more concentrated with fewer outliers. When the cumulative probability reaches approximately 0.4, the CDF curve of MFA-ANN is positioned to the left of the ANN method. Additionally, when the cumulative probability approaches 0.8, the MFA-ANN curve is positioned to the left of the LSTM method. Compared to both the ANN and LSTM models, the MFA-ANN model achieves a higher cumulative probability at lower error levels, indicating its superior prediction performance.

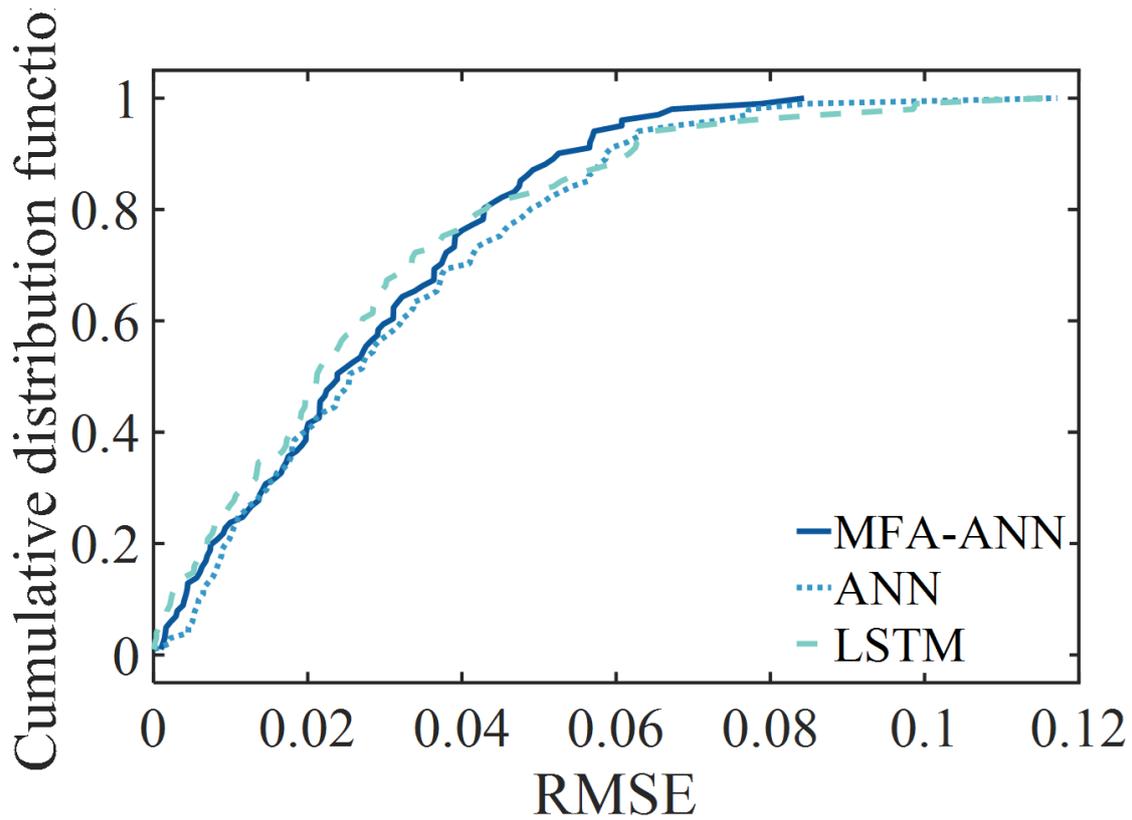

**Fig. 7.** Cumulative distribution of prediction errors of MFA-ANN model and ANN, LSTM model

**4.1.2 Comparison of prediction accuracy of different models**

To further validate the superiority of the proposed method, two common multivariate regression models, SVR and RF are selected as baseline methods for comparison. SVR captures the complex nonlinear relationship between inputs and outputs by applying the principle of interval maximization. RF, an ensemble learning method, enhances the model's robustness and generalization ability by aggregating multiple decision trees, making it particularly well-suited for handling high-dimensional and nonlinear data. To optimize these models, this study employs grid search combined with 5-fold cross-validation to identify the optimal hyperparameter combinations. For the SVR model, the key hyperparameters include the kernel function parameters and regularization coefficients. For the RF model, the hyperparameters include the number of trees, the maximum tree depth, and the minimum sample split.

As shown in **Fig. 8**, the RMSE of the MFA-ANN model is 0.0282, which significantly outperforms both the SVR (0.0378) and RF (0.0333) models. The prediction performance of the MFA-ANN model improves by 25.7% compared to SVR and by 15.6% compared to RF. Further analysis reveals that while RF performs better than SVR, the SVR model exhibits larger outliers in its predictions, which may significantly skew the overall error. To verify the statistical significance of these observed performance differences, a paired t-test is conducted.

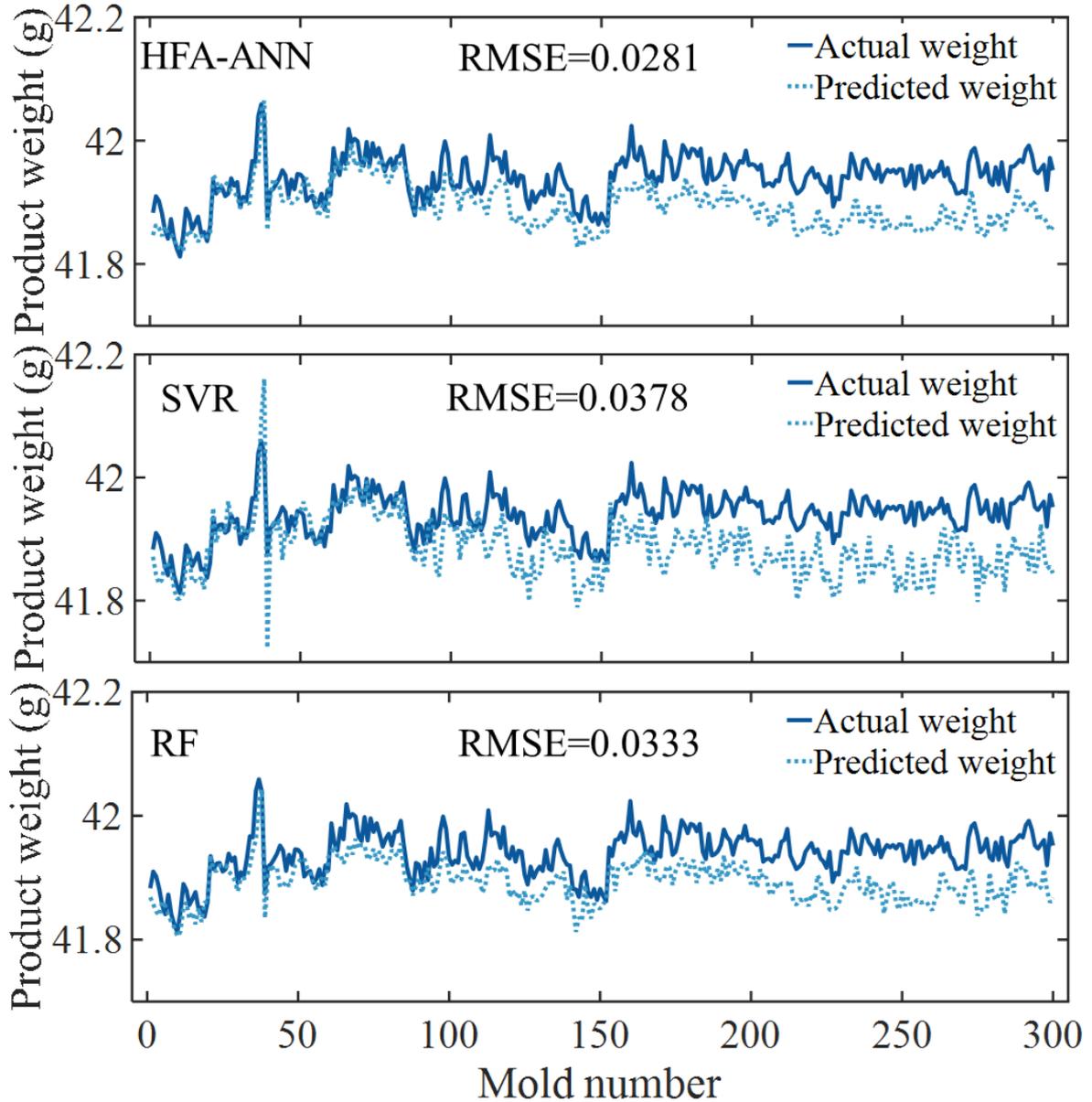

**Fig. 8.** Prediction results of different models: (a) MFA-ANN model; (b) SVR model; (c) RF model

Table 4 presents the p-values for the prediction performance differences between the various models. The p-values for the MFA-ANN model compared to the two baseline models are 0.03 and 7.81×10$^{-4}$ both of which are below the significance level of 0.05. This indicates that the proposed method significantly improves prediction performance. In contrast, the p-values for the SVR and RF models are 0.28, which is greater than the 0.05 significance level, suggesting that the difference in prediction performance between these two models is not statistically significant. This further indicates that the relatively poorer performance of the SVR model may be due to the influence of a few outliers, rather than a fundamental decrease in its overall prediction capability.

Table 4 P-values of prediction performance differences among different models

|         | MFA-ANN | SVR | RF |
| --- | --- | --- | --- |
| MFA-ANN | × | × | × |
| SVR | 0.03 | × | × |
| RF | 7.81×10$^{-4}$ | 0.28 | × |

**4.2 Ablation study results**

Ablation studies are a commonly used method for model analysis, allowing us to assess the contribution of each component to the overall performance by systematically removing individual components. To further evaluate the role of mixed feature modeling and the feature attention mechanism, this section conducts four sets of ablation studies. The specific experimental design is shown in Table 5. By comparing the model's performance under different configurations, we can gain a clearer understanding of how each component contributes to the final prediction accuracy. All hyperparameters of the model remain unchanged throughout the studies.

Table 5 Ablation study groups

| Group | Mixed feature modeling | Feature attention mechanism |
| --- | --- | --- |
| 1 | √ | √ |
| 2 | √ | × |
| 3 | × | √ |
| 4 | × | × |

**Fig. 9** illustrates the agreement between actual and predicted weights: the closer the points lie to the diagonal, the better the prediction. Among all models, MFA-ANN aligns most closely with the diagonal, confirming its superior accuracy. Quantitatively, introducing the mixed feature modeling strategy alone yields a 22.4% increase in accuracy, while the feature attention mechanism by itself contributes an 11.2 % gain. This underscores that mixed feature modeling is especially effective at capturing the time series dynamics of melt and temperature parameters. When both strategies are combined, the model's accuracy improves by 25.1%. Although this combined gain is slightly less than the boost from mixed-feature modeling alone, it reflects a clear synergistic effect, namely, enhancing overall feature representation while more effectively weighting their importance. To confirm the statistical significance of these improvements, we performed paired t-tests comparing each of the first three models against the baseline ANN. The calculated p-values are $4.52\times10^{-6}$, $5.88\times10^{-6}$, and $8.34\times10^{-3}$, all of which are below the preset significance level of 0.05. This indicates that the prediction performance of the first three models is significantly different from that of the baseline ANN model.

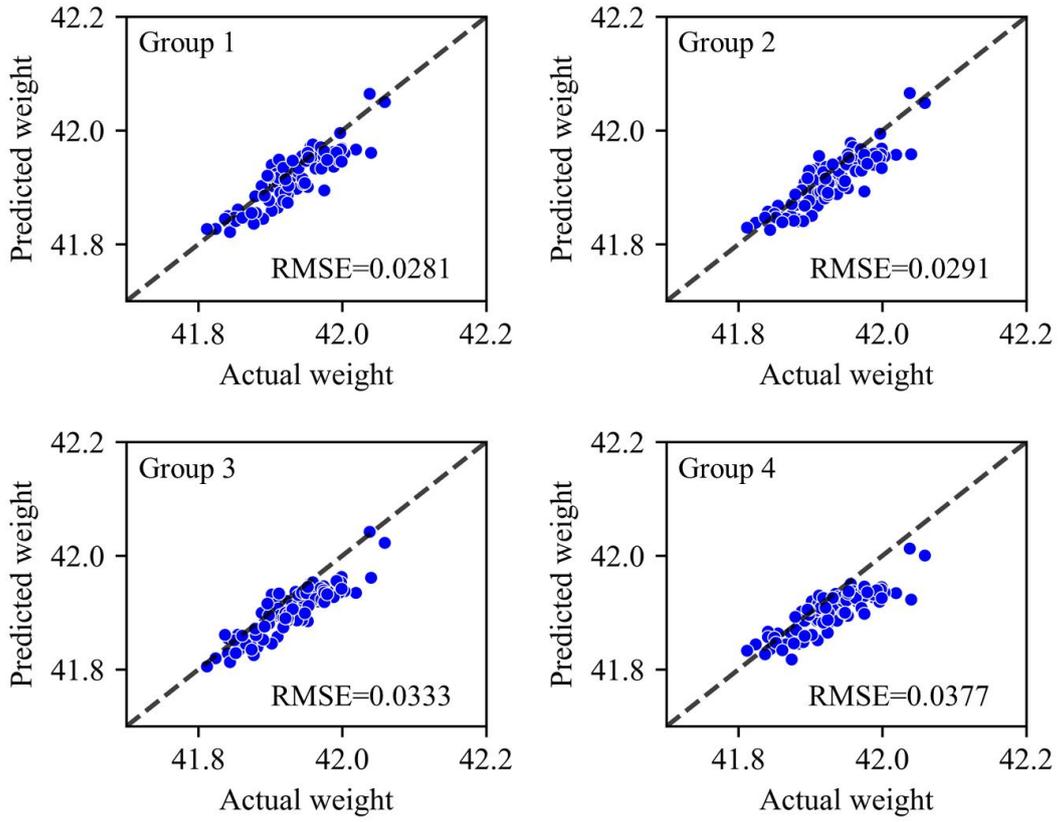

**Fig. 9.** Comparison of actual weight and predicted weight in ablation study

### 4.3 The impact of data accuracy on model prediction accuracy

The characteristic data of the injection molding machine is obtained through direct communication with the KEBA controller using the OPC UA protocol, ensuring high-precision molding process data. However, modern injection molding machines often offer the functionality of exporting characteristic data, such as Yizumi injection molding machines, which can export SPC data through a USB flash drive or by connecting to a data acquisition gateway. This exported data tends to lose precision, particularly for parameters like pressure and position, due to the conversion and rounding of the original data (referred to as low-precision data).

For example, the relationship between the melt starting point position parameter $S_{export}$ obtained through the export function and the melt starting point position $S_{KEBA}$ obtained through communication with the KEBA controller can be expressed as:

$$S_{export} =< \frac{4S_{KEBA}}{\pi D^2} > \quad (15)$$

where $<\cdot>$ represents the rounding operation on the data, and $D$ is the screw diameter of the injection molding machine.

To verify the impact of feature data precision on prediction accuracy, this study inputs both high-precision and low-precision data into the MFA-ANN model for training and prediction. In the experiment, the training and test sets remain unchanged, and the model hyperparameters are kept constant. **Fig. 10** presents the prediction error box plot for both types of data, visually highlighting the impact of data precision on prediction performance. The results demonstrate that when high-precision data is used, the RMSE of the model is 0.0281. However, when low-precision data is used, the RMSE increases to 0.0348, resulting in a 23.8% decrease in prediction accuracy. These results underscore the importance of high-precision feature data in improving model accuracy, suggesting that enhancing sensor precision could be an effective strategy to further optimize prediction performance.

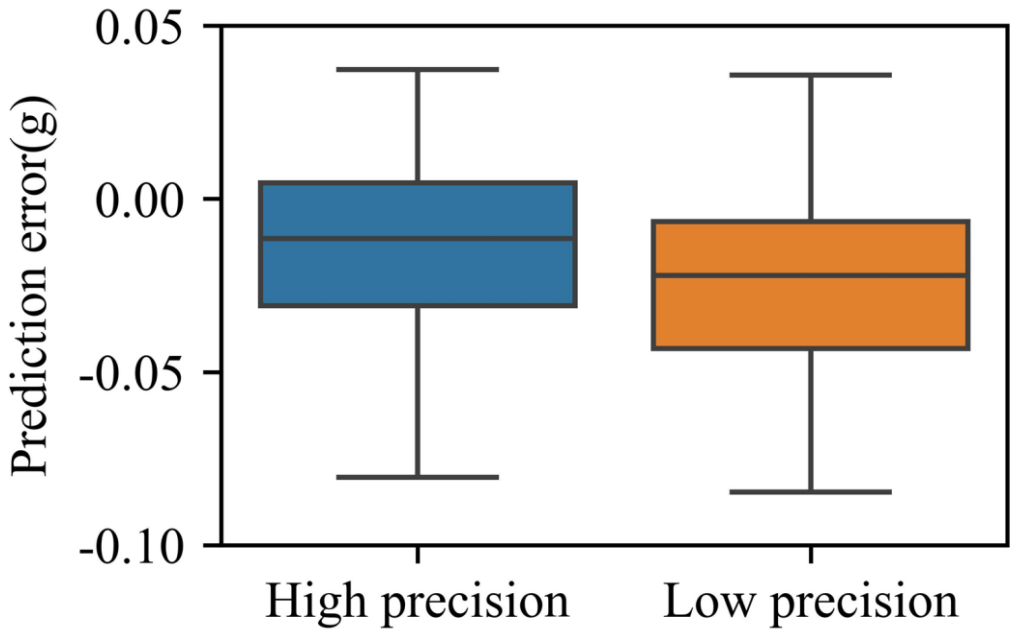

**Fig. 10.** Comparison of prediction results under different input data precision

## 5. Conclusions

In this study, an MFA-ANN model was proposed for online high-precision prediction of injection molding product weight, experimentally validated in an industrial production environment. The model achieves an RMSE value of 0.0281, showing a significant improvement in prediction accuracy. Compared to the ANN and LSTM baseline models with different feature input methods, the prediction accuracy increases by 25.1% and 23.0%, respectively. When compared to the SVR and RF baseline models, the prediction accuracy improves by 25.7% and 15.6%, respectively. These results demonstrate that the MFA-ANN model can more accurately predict the weight of injection molding products.

Additionally, the effects of the mixed feature modeling strategy and feature attention mechanism on prediction accuracy were analyzed. The mixed feature modeling strategy alone enhances prediction accuracy by 22.4%, while the feature attention mechanism improves it by 11.2%. When both strategies are combined, the prediction accuracy increases by 25.1%. These findings suggest that the mixed feature modeling strategy plays a more crucial role than the feature attention mechanism, and the synergy of both strategies significantly enhances feature representation and importance extraction. Finally, the impact of data accuracy on prediction performance was examined by training and testing the model with both high-precision and low-precision data. The RMSE of the model increased from 0.0281 to 0.0348, resulting in a 23.8% decrease in prediction accuracy. This underscores the importance of ensuring high-precision feature data collection to improve model accuracy.

**Declaration of competing interest**

The authors declare that they have no known competing financial interests or personal relationships that could have appeared to influence the work reported in this paper.

**Acknowledgments**

The project was supported by Natural Science Foundation of Xiamen (3502Z202571001), and Natural Science Foundation of Wuhan (2024040701010043).